\documentclass{article}

\usepackage[main, final]{neurips_2026}

\usepackage[utf8]{inputenc} % allow utf-8 input
\usepackage[T1]{fontenc}    % use 8-bit T1 fonts
\usepackage{hyperref}       % hyperlinks
\usepackage{url}            % simple URL typesetting
\usepackage{booktabs}       % professional-quality tables
\usepackage{amsfonts}       % blackboard math symbols
\usepackage{nicefrac}       % compact symbols for 1/2, etc.
\usepackage{microtype}      % microtypography
\usepackage{xcolor}         % colors
\usepackage{graphicx}
\usepackage{amsmath}
\usepackage{amssymb}

\usepackage[table]{xcolor}

\definecolor{externalrow}{HTML}{EDF4FA}
\definecolor{internalrow}{HTML}{FFF2E2}

\usepackage{tcolorbox}

\definecolor{promptbg}{RGB}{230, 230, 250}
\definecolor{promptborder}{RGB}{153, 153, 204}

\title{Do Agents Know When They Succeed? Calibrating Agent Confidence from Internal Representations}

\author{%
  Priyanka M.~Mammen$^1$ \qquad Emil Joswin$^2$ \qquad Srujananjali Medicherla$^2$ \\[4pt]
  $^1$UMass Amherst \qquad $^2$Independent Research \\
  \texttt{\small pmammen@umass.edu}
}

\begin{document}

\maketitle

\begin{abstract}
%Agentic systems are evolving rapidly and are now  capable of solving hard, multi-step reasoning problems. 

As agentic systems getting adopted rapidly in safety critical applications, it is vital to measure the confidence associated with the agentic actions. In comparison to the traditional machine learning systems, agentic workflows have complex failure modes  with  planning, tool invocation and dynamic environment interactions.   In this paper, we investigate  whether  model's internal representations provide stronger signals of eventual task
success in multi-turn agentic setups. We introduce two complementary methods: \emph{Latent Trajectory
Dynamics} \textbf{(LTD)}, which summarizes changes in  residual-stream representations across an interaction trajectory, and the \emph{Action Representation Probe} \textbf{(ARP)},
which predicts success from representations formed at action decisions. Across three interactive benchmarks (Bash, SQL, Python) and three model families (Qwen-14B, Qwen-7B, DeepSeek-6.7B), our methods consistently outperform surface-level generation and sequence-based calibration baselines providing a zero-overhead reliability monitor that requires neither prompt alterations nor multi-sample rollouts.

% In this paper, we investigate whether mechanistically derived confidence signals can provide fine-grained confidence signals through multi-turn agentic interactions.  We conducted experiments using intercode-bash, python and sql with multi-turn setup on models Qwen-14B
% Qwen-7B, DeepSeek-6.7B, we find that signals derived from model's internal states consistently outperform external confidence signals.
\end{abstract}

\section{Introduction}

Autonomous agentic systems are getting popular and increasingly being deployed across safety-critical domains including software-engineering tasks, trading, finance, clinical decision making, and robotic systems. Therefore, reliability and failure detection is essential for  its deployment. During real-time execution, agents operate under no ground truth label, and so the general strategy is to rely on an evaluation step to determine intervention or implement any fallback policies.

Traditional confidence estimation techniques designed for single-turn generation like temperature scaling, output token semantic entropy, verbalized confidence \cite{zhao2026wired,tian2023just,xu2025language} etc struggle in agentic workflows.  Unlike typical text generation tasks, in agentic settings, models do not work in isolation, they operate within a harness that orchestrate multi-turn planning, tools, and feedback from external environment over long-horizon tasks. Hence, due to the nature of complex execution of agents in long-horizon tasks, these static system techniques won't work for multi-turn. \cite{oh2026uncertainty}. Recent works such as \cite{duan2025uprop, zhao2025uncertainty} came up with a confidence estimation techniques which consider error propagation across steps in a reasoning trajectory and another work \cite{zhang2026agentic} proposed an agentic confidence calibration from trajectory derived signals. As these methods leverage only output level signals, they might fail to capture the model inner states preceding the model failures. \cite{azaria2023internal}  has shown that model inner states can better indicate the model veracity than model outputs.  

 % In light of this a recent work proposed an agentic confidence calibration from trajectory derived signals. Whats the main problem with this paper? \cite{ stoisser2025towards}, perplexity-based confidence estimation for model fine-tuning. RL based framework to improve verbalized confidence \cite{xuan2026confidence}

Motivated by this insight, we  turn towards mechanistic approaches to derive granular level model confidence signals. 
Mechanistic interpretability related works in agentic settings are mostly single-turn based and used for  tool-need decision making \cite{liu-etal-2024-uncertainty,subramani2025mice, healy2026internal, tatsat2026beyond} in reasoning tasks. Building upon this foundational work, we investigate different mechanistic approaches to derive confidence in multi-turn agentic setups.

Our main contributions are as follows:
\begin{itemize}
    \item We introduce two novel internal confidence extraction frameworks operating directly on the agent's hidden activations: \textbf{Latent Trajectory Dynamics (LTD)}, which quantifies the geometric stability and drift of the agent's execution path, and the \textbf{Action Representation Probe (ARP)}, which decodes task success directly from the action-span residual states.
    \item We benchmark surface token logprobs, trajectory dynamics, and internal representations under a leak-free cross-validation protocol and show that across InterCode-Bash, SQL, and Python on three open-weight coding models (Qwen-14B, Qwen-7B, DeepSeek-6.7B), internal methods consistently dominate surface baselines, establishing internal state probing as a viable zero-overhead reliability safeguard for deployed agents.
\end{itemize}

\section{Methodology}
\label{sec:methodology}

\subsection{Problem Formulation}
\label{sec:problem-formulation}

For task $i$, a coding agent produces a multi-step trajectory
$\tau_i=\{(g_{i,t},o_{i,t})\}_{t=1}^{T_i}$, where $g_{i,t}$ is the generation
at step $t$, $o_{i,t}$ is the resulting observation, and
$y_i\in\{0,1\}$ indicates task success. We seek a trajectory-level confidence
estimate
\[
\hat c_i \approx P(y_i=1\mid\tau_i).
\]

For each generated token, we retain its log-probability $\ell_{i,t,j}$ and
recover its probability as $p_{i,t,j}=\exp(\ell_{i,t,j})$. We also recover
residual-stream states $h^{(\ell)}_{i,t,j}$ by teacher-forcing the stored
trajectory through the generating model which, in live deployment are computed during generation with zero replay overhead; see Appendix~\ref{app:deployment}. Calibrated Logprob  and Holistic Trajectory Calibration (HTC) \cite{zhang2026agentic} estimate
confidence from observable trajectory information, whereas LTD and ARP use
these internal representations. More details about external confidence baselines are provided in Appendix~\ref{app:external-signals}.

\subsection{Internal Representation Signals}
\label{sec:internal-signals}

\subsubsection{Latent Trajectory Dynamics}
\label{sec:ltd}

Latent Trajectory Dynamics (LTD) describes how the model's internal state
changes over an agent trajectory. We teacher-force each stored generation and
retain residual-stream states at semantically meaningful endpoints
corresponding to the task observation, reasoning, action, and subsequent
feedback. From adjacent states $h_t$ and $h_{t+1}$, we measure cosine
displacement
\begin{equation}
    d_{\mathrm{cos}}\left(h_t,h_{t+1}\right)
    =
    1 -
    \frac{
        h_t^\top h_{t+1}
    }{
        \lVert h_t \rVert_2
        \lVert h_{t+1} \rVert_2
    },
\end{equation}
and relative displacement
\begin{equation}
    d_{\mathrm{rel}}\left(h_t,h_{t+1}\right)
    =
    \frac{
        \lVert h_{t+1} - h_t \rVert_2
    }{
        \lVert h_t \rVert_2 + \epsilon
    }.
\end{equation}

We summarize these quantities over reasoning, commitment, action, and feedback
transitions using their means, final values, variability, and trends. We
additionally measure path efficiency and include indicators for unavailable
transition types. Full set of features are present in Appendix \ref{app:features}. This produces
$x_i^{\mathrm{LTD}} \in \mathbb{R}^{28}$. To isolate the information in these
internal dynamics while matching HTC's predictor class, we fit an
L2-regularized logistic model.
%
% \begin{equation}
%     \hat{c}_i
%     =
%     \sigma\left(
%         w^\top x_i^{\mathrm{LTD}} + b
%     \right).
% \end{equation}

\subsubsection{Action Representation Probe}
\label{sec:arp}

The Action Representation Probe (ARP) asks whether the internal representation
at action decisions directly encodes eventual task success. Let $a_{i,t}$
denote the final-layer residual state at the endpoint of the action span at
step $t$. We exclude terminal submit actions and mean-pool the remaining action
states:
\begin{equation}
    \bar{a}_i
    =
    \frac{1}{\lvert \mathcal{A}_i \rvert}
    \sum_{t \in \mathcal{A}_i}
    a_{i,t}
\end{equation}

Within each training fold, we standardize these episode representations and
project them onto their first 64 principal components:
\begin{equation}
    u_i
    =
    \operatorname{PCA}_{64}
    \left(
        \operatorname{Standardize}\left(\bar{a}_i\right)
    \right).
\end{equation}
An L2-regularized logistic probe predicts success from the projected
representation.
%
% \begin{equation}
%     \tilde{c}_i
%     =
%     \sigma\left(w^\top u_i + b\right).
% \end{equation}
%
A monotonic Platt map subsequently converts this score into the reported
confidence. Standardization, PCA, probe fitting, and probability calibration
are all performed without access to the outer held-out fold.

%\subsection{Evaluation Protocol}

% \subsection{Probing Experiments}

% Let episode $i$ have turns $t = 1, \dots, T_i$, and let $\mathbf{h}_{i,t}^{(L)} \in \mathbb{R}^d$ be the residual-stream state at the action-span end token of turn $t$ at the final layer $L$ ($d = \text{hidden size}$). The episode feature is the mean-pooled action state:
% \[
% \mathbf{x}_i = \frac{1}{T_i} \sum_{t=1}^{T_i} \mathbf{h}_{i,t}^{(L)}
% \]
% It is standardized per dimension (training mean $\boldsymbol{\mu}$, training standard deviation $\mathbf{s}$) and projected onto the top $k = 64$ training principal components:
% \[
% \mathbf{z}_i = \frac{\mathbf{x}_i - \boldsymbol{\mu}}{\mathbf{s}}
% \]
% \[
% \mathbf{u}_i = \mathbf{W}_{\text{PCA}}^\top \mathbf{z}_i \in \mathbb{R}^k
% \]
% with $\boldsymbol{\mu}$, $\mathbf{s}$, and $\mathbf{W}_{\text{PCA}}$ estimated on the training folds only. A logistic probe gives an uncalibrated score, which a monotone calibration map $g(\cdot)$ (isotonic or Platt, fit on held-out training data) converts into the reported confidence:
% \[
% \tilde{p}_i = \sigma(\mathbf{w}^\top \mathbf{u}_i + b)
% \]
% \[
% \hat{c}_i = g(\tilde{p}_i) \in [0, 1]
% \]
% where $\sigma(z) = \frac{1}{1 + e^{-z}}$ is the logistic function. $\hat{c}_i$ is the model's estimated probability that episode $i$ is correct.

\section{Experiments}

\subsection{Experimental Setup}

\textbf{Models and Agentic Framework:}
We used three models for our evaluations -
Qwen2.5-Coder-7B-Instruct, Qwen2.5-Coder-14B-Instruct-AWQ, and DeepSeek-Coder-6.7B-Instruct, on a single A100 GPU. All models are served locally using vLLM \citep{kwon2023efficient} via its OpenAI-compatible API, which exposes per-token log-probabilities. The prompts samples are provided in Appendix\ref{app:system_prompts}. All reported results use greedy decoding (temperature = 0). Later we teacher-force the generated trajectories on the same offline model to obtain the residual stream states. For each model and environment, all methods are evaluated on the same task
population using identical frozen outer folds, producing one out-of-fold
prediction per trajectory. More details are provided in Appendix~\ref{app:calibration-protocol}.

% Reference:The agentic loop is implemented as a custom multi-turn reasoning loop. Both model families support native structured tool calling, Qwen3 models emit $<tool\_call>$ JSON blocks as part of their output, while Hermes-3 uses the same format parsed via vLLM's built-in Hermes tool-call parser. On each turn, the model may invoke one of three registered tools: a $calculator$ for arithmetic (backed by SymPy), a $solve$ function for symbolic equations, or a lightweight $think$ step for explicit intermediate reasoning. When hint injection is enabled, an additional $retrieve\_hint$ tool is made available. The loop continues until the model produces a final answer without a tool call. We implement this loop directly rather than through a high-level agent framework, so that per-token log-probabilities and raw reasoning trajectories remain fully accessible for downstream confidence estimation.

\textbf{Datasets:} 
We ground our investigation in interactive coding benchmarks as they afford objective, deterministic ground-truth verification via programmatic unit tests and execution harnesses completely eliminating LLM-as-a-judge evaluation noise while also requiring multi-turn interactive problem-solving across diverse syntactic and semantic modalities.
We used three types of interactive coding tasks from intercode-bench \cite{yang2023intercode} which consists of python (MBPP), sql (SPIDER) and bash  tasks.  
More details are given in Appendix \ref{app:benchmarks}.

\textbf{Baselines and Eval Metrics:} We benchmark our approach against the external confidence signals - HTC(\cite{zhang2026agentic}) and calibrated log prob. 
We use the three standard confidence calibration scores. AUROC (Area Under Receiver Operating Characteristic) measures the ability to distinguish correct and incorrect reasoning trajectories;
 ECE measures the difference between a models predicted probabilities and the observed accuracy \cite{guo2017calibration}; and Brier Score measures the  mean squared difference between predicted probabilities and actual binary outcome \cite{glenn1950verification}. For a good calibration metric, we expect a higher AUROC score with a lower Brier score and  a lower ECE score.

\subsection{Calibration and Performance Results}
Table~\ref{tab:confidence-results} summarizes out-of-fold discrimination (AUROC) and calibration metrics (Brier score, ECE) across all three evaluation environments and model architectures.  Our empirical findings demonstrate three core trends:
\begin{enumerate}
    \item \textbf{Token confidence is an insufficient external signal.}
    Calibrated Logprob provides the weakest discrimination, indicating
    that aggregate output probability alone does not reliably capture  trajectory-level success.

    \item \textbf{Trajectory structure strengthens external confidence
    estimation.} HTC substantially improves over Cal Logprob in most
    settings, showing that confidence dynamics across an interaction are more
    informative than a single aggregate probability.

    \item \textbf{Internal representations provide confidence signals beyond
    observable outputs.} LTD and ARP outperform the external baselines on
    AUROC or Brier score in every model--environment setting, with an internal
    method achieving the best AUROC and Brier score in all nine settings.
    Together, these results show that both trajectory-level dynamics and
    action-level representations expose information about eventual success
    that is not captured by output probabilities alone.
\end{enumerate}

\begin{table*}[t]
\centering
\caption{
Confidence evaluation across calibration paradigms and model families. Methods progress from surface outputs (\textbf{Logprob (Cal.)}) to sequence dynamics (\textbf{HTC}) and internal representation mechanisms (\textbf{LTD}, \textbf{ARP}). All metrics use 5-fold nested cross-validation on frozen protocols.
}
\vspace{10px}
\label{tab:confidence-results}

\begingroup
\small
\setlength{\tabcolsep}{3.5pt}
\renewcommand{\arraystretch}{1.02}

% ------------------------------------------------------------------------------
% Qwen 14B
% ------------------------------------------------------------------------------
\begin{minipage}{\textwidth}
\centering
\textbf{(a) Qwen2.5-Coder-14B-Instruct-AWQ}\\[-1pt]
\end{minipage}

\smallskip

\begin{tabular*}{\textwidth}{
    @{\extracolsep{\fill}}
    l
    ccc
    ccc
    ccc
    @{}
}
\toprule
& \multicolumn{3}{c}{\textbf{Bash} (33.5\%)}
& \multicolumn{3}{c}{\textbf{SQL} (71.3\%)}
& \multicolumn{3}{c}{\textbf{Python} (51.0\%)} \\
\cmidrule(lr){2-4}
\cmidrule(lr){5-7}
\cmidrule(lr){8-10}
\textbf{Method}
& AUROC $\uparrow$ & Brier $\downarrow$ & ECE $\downarrow$
& AUROC $\uparrow$ & Brier $\downarrow$ & ECE $\downarrow$
& AUROC $\uparrow$ & Brier $\downarrow$ & ECE $\downarrow$ \\
\midrule
\rowcolor{externalrow}
Logprob (Cal.)
& 0.627 & 0.218 & 0.112
& 0.624 & 0.197 & \textbf{0.018}
& 0.555 & 0.247 & \textbf{0.021} \\
\rowcolor{externalrow}
HTC
& 0.743 & 0.196 & 0.095
& 0.743 & 0.176 & 0.052
& 0.655 & 0.231 & 0.044 \\
\rowcolor{internalrow}
LTD
& 0.761 & 0.183 & \textbf{0.069}
& 0.771 & 0.169 & 0.044
& 0.646 & 0.234 & 0.025 \\
\rowcolor{internalrow}
ARP
& \textbf{0.814} & \textbf{0.162} & 0.070
& \textbf{0.842} & \textbf{0.144} & 0.055
& \textbf{0.711} & \textbf{0.216} & 0.028 \\
\bottomrule
\end{tabular*}

\medskip

% ------------------------------------------------------------------------------
% Qwen 7B
% ------------------------------------------------------------------------------
\begin{minipage}{\textwidth}
\centering
\textbf{(b) Qwen2.5-Coder-7B-Instruct}\\[-1pt]
\end{minipage}

\smallskip

\begin{tabular*}{\textwidth}{
    @{\extracolsep{\fill}}
    l
    ccc
    ccc
    ccc
    @{}
}
\toprule
& \multicolumn{3}{c}{\textbf{Bash} (30.3\%)}
& \multicolumn{3}{c}{\textbf{SQL} (67.4\%)}
& \multicolumn{3}{c}{\textbf{Python} (47.3\%)}  \\
\cmidrule(lr){2-4}
\cmidrule(lr){5-7}
\cmidrule(lr){8-10}
\textbf{Method}
& AUROC $\uparrow$ & Brier $\downarrow$ & ECE $\downarrow$
& AUROC $\uparrow$ & Brier $\downarrow$ & ECE $\downarrow$
& AUROC $\uparrow$ & Brier $\downarrow$ & ECE $\downarrow$ \\
\midrule
\rowcolor{externalrow}
Logprob (Cal.)
& 0.510 & 0.216 & 0.052
& 0.702 & 0.197 & 0.028
& 0.509 & 0.249 & 0.047 \\
\rowcolor{externalrow}
HTC
& 0.575 & 0.221 & 0.107
& 0.800 & 0.164 & \textbf{0.023}
& 0.684 & 0.222 & \textbf{0.020} \\
\rowcolor{internalrow}
LTD
& 0.629 & 0.212 & 0.109
& 0.786 & 0.171 & 0.057
& 0.659 & 0.229 & 0.028 \\
\rowcolor{internalrow}
ARP
& \textbf{0.704} & \textbf{0.190} & \textbf{0.048}
& \textbf{0.837} & \textbf{0.151} & 0.038
& \textbf{0.718} & \textbf{0.214} & 0.024 \\
\bottomrule
\end{tabular*}

\medskip

% ------------------------------------------------------------------------------
% DeepSeek 6.7B
% ------------------------------------------------------------------------------
\begin{minipage}{\textwidth}
\centering
\textbf{(c) DeepSeek-Coder-6.7B-Instruct}\\[-1pt]
\end{minipage}

\smallskip

\begin{tabular*}{\textwidth}{
    @{\extracolsep{\fill}}
    l
    ccc
    ccc
    ccc
    @{}
}
\toprule
& \multicolumn{3}{c}{\textbf{Bash} (22.5\%)}
& \multicolumn{3}{c}{\textbf{SQL} (50.0\%)}
& \multicolumn{3}{c}{\textbf{Python} (35.6\%)} \\
\cmidrule(lr){2-4}
\cmidrule(lr){5-7}
\cmidrule(lr){8-10}
\textbf{Method}
& AUROC $\uparrow$ & Brier $\downarrow$ & ECE $\downarrow$
& AUROC $\uparrow$ & Brier $\downarrow$ & ECE $\downarrow$
& AUROC $\uparrow$ & Brier $\downarrow$ & ECE $\downarrow$ \\
\midrule
\rowcolor{externalrow}
Logprob (Cal.)
& 0.482 & 0.175 & \textbf{0.006}
& 0.454 & 0.251 & 0.042
& 0.520 & 0.230 & \textbf{0.002} \\
\rowcolor{externalrow}
HTC
& 0.698 & 0.165 & 0.075
& 0.726 & 0.211 & 0.041
& 0.645 & 0.218 & 0.038 \\
\rowcolor{internalrow}
LTD
& \textbf{0.727} & \textbf{0.162} & 0.093
& 0.624 & 0.239 & 0.039
& 0.670 & 0.211 & 0.037 \\
\rowcolor{internalrow}
ARP
& 0.650 & 0.167 & 0.042
& \textbf{0.788} & \textbf{0.187} & \textbf{0.029}
& \textbf{0.784} & \textbf{0.179} & 0.040 \\
\bottomrule
\end{tabular*}

\endgroup
\end{table*}

\section{Related Work}

 The early  approaches on agents directly adapt from static language models including temperature scaling \cite{guo2017calibration},  semantic entropy of predicted tokens \cite{kuhn2023semantic}, and, verbalized confidence\cite{tian2023just}. All these approaches focus on single-turn outputs and did not handle multiple steps in a trajectory together. Works such as \cite{duan2025uprop, zhao2025uncertainty} take into account how uncertainty is propagated  across different steps in  the agentic reasoning trajectory and a trajectory derived classifier proposed by \cite{zhang2026agentic}  
The other category of work look at different mechanistic signals - execution traces and embedding-based probes (\cite{liu-etal-2024-uncertainty}),  unembedding similarity signals between the layers \cite{subramani2025mice} for guiding tool use and interpretability using different steps in the  chain-of-thought  reasoning\cite{sun2026llm}. Our work tries to bridge gap between  the current mechanistic interpretability  and long-horizon agentic tasks.

%Many works \cite{subramani2025mice, tatsat2026beyond,liu-etal-2024-uncertainty} have empirically shown that models tend to be overconfident especially given harder or out of the distribution inputs. Recent work such as X \cite{leng2025taming, xuan2026confidence} proposed a rl-training framework for agentic calibration to improve the verbalized confidence of the models.

\section{Conclusion}
\label{sec:conclusion}

In this work, we demonstrate that passive introspection of an LLM agent's internal residual stream provides a far more reliable confidence signal than surface generation text or sequence-level attention statistics. Across three interactive benchmarks (Bash, SQL, Python) and three model families (Qwen-14B, Qwen-7B, DeepSeek-6.7B), our proposed internal frameworks i)Latent Trajectory Dynamics \textbf{(LTD)} and ii) Action Representation Probing \textbf{(ARP)} consistently outperform state-of-the-art token-level calibration methods. Because our approach operates entirely on representations computed during standard forward passes with zero prompt modifications or multi-rollout sampling, it offers an efficient, passive reliability safeguard for autonomous agents deployed in real-world environments. These results motivate internal-state monitoring as a promising foundation for detecting unreliable agent behavior and enabling confidence-aware intervention during deployment.

\bibliographystyle{plainnat} % or neurips_2026 if specified
\bibliography{references}

@inproceedings{zhao2025uncertainty,
  title={Uncertainty propagation on llm agent},
  author={Zhao, Qiwei and Li, Dong and Liu, Yanchi and Cheng, Wei and Sun, Yiyou and Oishi, Mika and Osaki, Takao and Matsuda, Katsushi and Yao, Huaxiu and Zhao, Chen and others},
  booktitle={Proceedings of the 63rd Annual Meeting of the Association for Computational Linguistics (Volume 1: Long Papers)},
  pages={6064--6073},
  year={2025}
}

@article{tatsat2026beyond,
  title={Beyond the Black Box: Interpretability of Agentic AI Tool Use},
  author={Tatsat, Hariom and Shater, Ariye},
  journal={arXiv preprint arXiv:2605.06890},
  year={2026}
}

@inproceedings{subramani2025mice,
  title={MICE for CATs: Model-internal confidence estimation for calibrating agents with tools},
  author={Subramani, Nishant and Eisner, Jason and Svegliato, Justin and Van Durme, Benjamin and Su, Yu and Thomson, Sam},
  booktitle={Proceedings of the 2025 Conference of the Nations of the Americas Chapter of the Association for Computational Linguistics: Human Language Technologies (Volume 1: Long Papers)},
  pages={12362--12375},
  year={2025}
}

@inproceedings{liu-etal-2024-uncertainty,
    title = "Uncertainty Calibration for Tool-Using Language Agents",
    author = "Liu, Hao  and
      Dou, Zi-Yi  and
      Wang, Yixin  and
      Peng, Nanyun  and
      Yue, Yisong",
    editor = "Al-Onaizan, Yaser  and
      Bansal, Mohit  and
      Chen, Yun-Nung",
    booktitle = "Findings of the Association for Computational Linguistics: EMNLP 2024",
    month = nov,
    year = "2024",
    address = "Miami, Florida, USA",
    publisher = "Association for Computational Linguistics",
    url = "https://aclanthology.org/2024.findings-emnlp.978/",
    doi = "10.18653/v1/2024.findings-emnlp.978",
    pages = "16781--16805"
}

@article{zhang2026agentic,
  title={Agentic confidence calibration},
  author={Zhang, Jiaxin and Xiong, Caiming and Wu, Chien-Sheng},
  journal={arXiv preprint arXiv:2601.15778},
  year={2026}
}

@article{duan2025uprop,
  title={Uprop: Investigating the uncertainty propagation of llms in multi-step agentic decision-making},
  author={Duan, Jinhao and Diffenderfer, James and Madireddy, Sandeep and Chen, Tianlong and Kailkhura, Bhavya and Xu, Kaidi},
  journal={arXiv preprint arXiv:2506.17419},
  year={2025}
}

@inproceedings{sun2026llm,
  title={Llm reasoning as trajectories: Step-specific representation geometry and correctness signals},
  author={Sun, Lihao and Dong, Hang and Qiao, Bo and Lin, Qingwei and Zhang, Dongmei and Rajmohan, Saravan},
  booktitle={Proceedings of the 64th Annual Meeting of the Association for Computational Linguistics (Volume 1: Long Papers)},
  pages={26872--26887},
  year={2026}
}

@article{zhao2026wired,
  title={Wired for overconfidence: A mechanistic perspective on inflated verbalized confidence in llms},
  author={Zhao, Tianyi and He, Yinhan and Zheng, Wendy and Zhang, Yujie and Chen, Chen},
  journal={arXiv preprint arXiv:2604.01457},
  year={2026}
}

@inproceedings{tian2023just,
  title={Just ask for calibration: Strategies for eliciting calibrated confidence scores from language models fine-tuned with human feedback},
  author={Tian, Katherine and Mitchell, Eric and Zhou, Allan and Sharma, Archit and Rafailov, Rafael and Yao, Huaxiu and Finn, Chelsea and Manning, Christopher D},
  booktitle={Proceedings of the 2023 Conference on Empirical Methods in Natural Language Processing},
  pages={5433--5442},
  year={2023}
}

@inproceedings{xu2025language,
  title={Do language models mirror human confidence? exploring psychological insights to address overconfidence in LLMs},
  author={Xu, Chenjun and Wen, Bingbing and Han, Bin and Wolfe, Robert and Wang, Lucy Lu and Howe, Bill},
  booktitle={Findings of the Association for Computational Linguistics: ACL 2025},
  pages={25655--25672},
  year={2025}
}

@article{kuhn2023semantic,
  title={Semantic uncertainty: Linguistic invariances for uncertainty estimation in natural language generation},
  author={Kuhn, Lorenz and Gal, Yarin and Farquhar, Sebastian},
  journal={arXiv preprint arXiv:2302.09664},
  year={2023}
}

@inproceedings{guo2017calibration,
  title={On calibration of modern neural networks},
  author={Guo, Chuan and Pleiss, Geoff and Sun, Yu and Weinberger, Kilian Q},
  booktitle={International conference on machine learning},
  pages={1321--1330},
  year={2017},
  organization={PMLR}
}

@article{yang2023intercode,
  title={Intercode: Standardizing and benchmarking interactive coding with execution feedback},
  author={Yang, John and Prabhakar, Akshara and Narasimhan, Karthik and Yao, Shunyu},
  journal={Advances in Neural Information Processing Systems},
  volume={36},
  pages={23826--23854},
  year={2023}
}

@article{healy2026internal,
  title={Internal representations as indicators of hallucinations in agent tool selection},
  author={Healy, Kait and Srinivasan, Bharathi and Madathil, Visakh and Wu, Jing},
  journal={arXiv preprint arXiv:2601.05214},
  year={2026}
}

@inproceedings{azaria2023internal,
  title={The internal state of an LLM knows when it’s lying},
  author={Azaria, Amos and Mitchell, Tom},
  booktitle={Findings of the Association for Computational Linguistics: EMNLP 2023},
  pages={967--976},
  year={2023}
}

@article{glenn1950verification,
  title={Verification of forecasts expressed in terms of probability},
  author={Glenn, W Brier and others},
  journal={Monthly weather review},
  volume={78},
  number={1},
  pages={1--3},
  year={1950}
}

@inproceedings{kwon2023efficient,
  title={Efficient memory management for large language model serving with pagedattention},
  author={Kwon, Woosuk and Li, Zhuohan and Zhuang, Siyuan and Sheng, Ying and Zheng, Lianmin and Yu, Cody Hao and Gonzalez, Joseph and Zhang, Hao and Stoica, Ion},
  booktitle={Proceedings of the 29th symposium on operating systems principles},
  pages={611--626},
  year={2023}
}

@inproceedings{oh2026uncertainty,
  title={Uncertainty quantification in llm agents: Foundations, emerging challenges, and opportunities},
  author={Oh, Changdae and Park, Seongheon and Kim, To Eun and Li, Jiatong and Li, Wendi and Yeh, Samuel and Du, Sean and Hassani, Hamed and Bogdan, Paul and Song, Dawn and others},
  booktitle={Proceedings of the 64th Annual Meeting of the Association for Computational Linguistics (Volume 1: Long Papers)},
  pages={16219--16250},
  year={2026}
}

@article{austin2021program,
  title={Program Synthesis with Large Language Models},
  author={Austin, Jacob and others},
  journal={arXiv preprint arXiv:2108.07732},
  year={2021}
}

@inproceedings{yu2018spider,
  title={Spider: A Large-Scale Human-Labeled Dataset for Complex and Cross-Domain Semantic Parsing and Text-to-SQL Task},
  author={Yu, Tao and Zhang, Rui and Yang, Kai and Yasunaga, Michihiro and Wang, Dongxu and Li, Zifan and Ma, James and Li, Irene and Yao, Qingning and Roman, Shanelle and others},
  booktitle={Proceedings of the 2018 Conference on Empirical Methods in Natural Language Processing},
  pages={3911--3921},
  year={2018}
}

\appendix

\section{Appendix}

\section{InterCode System Prompts}
\label{app:system_prompts}

\begin{tcolorbox}[colback=promptbg, colframe=promptborder, boxrule=2pt, arc=0pt, left=20pt, right=20pt, top=15pt, bottom=15pt]
\subsubsection*{\large BASH}

\textit{You are a helpful assistant solving tasks in a Bash shell.
You will be given a task. Solve it by running one shell command at a time.}

Respond in EXACTLY this format each turn:
\texttt{THOUGHT:} \texttt{<your reasoning>}
\texttt{ACTION:} \texttt{<a single shell command, OR the literal word submit if the task is complete>}

Rules:
\begin{itemize}
  \item Only one command per ACTION line.
  \item When you believe the task is complete, respond with ACTION: submit
  \item Do not use interactive commands (vi, nano, etc).
\end{itemize}

% \vspace{3pt}

\end{tcolorbox}

% \vspace{20pt}

\begin{tcolorbox}[colback=promptbg, colframe=promptborder, boxrule=2pt, arc=0pt, left=20pt, right=20pt, top=15pt, bottom=15pt]
\subsubsection*{\large PYTHON (MBPP)}

\textit{You are a helpful assistant solving Python function-writing task.
You can execute Python code in a persistent interpreter, inspect its output or errors, revise your implementation, and finally submit a function for testing.}

Respond in exactly this format:
\texttt{THOUGHT:} \texttt{<brief reasoning>}
\texttt{ACTION:}
\texttt{```python}
\texttt{<valid Python code>}
\texttt{```}

To finish, use a single-line action instead:
\texttt{ACTION: submit <function\_name>}

Rules:
\begin{itemize}
  \item Execute complete Python statements or complete function definitions.
  \item Do not include text after the action.
  \item State persists across execution turns and resets between tasks.
  \item Submit only a function that has already been defined successfully.
\end{itemize}

% \vspace{3pt}

\end{tcolorbox}

% \vspace{20pt}

\begin{tcolorbox}[colback=promptbg, colframe=promptborder, boxrule=2pt, arc=0pt, left=20pt, right=20pt, top=15pt, bottom=15pt]
\subsubsection*{\large SQL (SPIDER)}

\textit{You are a helpful assistant solving tasks in a MySQL database.
You will be given a question and the relevant database schema. Solve it by running one SQL statement at a time. You may inspect the database using SQL such as SHOW TABLES and DESCRIBE table\_name.}

Respond in EXACTLY this format each turn:
\texttt{THOUGHT:} \texttt{<your reasoning>}
\texttt{ACTION:} \texttt{<a single SQL statement, OR the literal word submit if the latest query result answers the question>}

Rules:
\begin{itemize}
  \item Put exactly one SQL statement on the ACTION line.
  \item Do not include Markdown code fences.
  \item Do not execute data-changing or administrative statements.
  \item When the latest query result is your final answer, respond with ACTION: submit.
\end{itemize}

\vspace{8pt}

\end{tcolorbox}

% \begin{figure}[h]
%     \centering
%     \includegraphics[width=\linewidth]{images/all_prompts.png}
%     \caption{Prompts used for the intercode tasks}
%     \label{fig:placeholder}
% \end{figure}

\section{Benchmark Environments \& Trajectory Statistics}
\label{app:benchmarks}

We evaluate all calibration and probing methods across three distinct interactive agent environments from the InterCode benchmark suite \citep{yang2023intercode}:
\begin{enumerate}
    \item \textbf{InterCode-Bash:} Evaluates multi-turn Linux shell command execution for operating system administration, file manipulation, and process orchestration. Tasks require executing bash commands, parsing stdout and stderr feedback, and iteratively recovering from runtime execution errors.
    \item \textbf{InterCode-SQL:} Tests natural-language-to-SQL generation and multi-turn database querying across Spider schemas \citep{yu2018spider}. The agent inspects schema definitions, executes interactive queries, receives table result sets or database syntax errors, and iteratively refines SQL queries until submission.
    \item \textbf{InterCode-Python:} Evaluates interactive code synthesis and unit-test execution on MBPP programming challenges \citep{austin2021program}. The agent synthesizes Python functions, executes test assertions in a live REPL, and debugs tracebacks.
\end{enumerate}

\begin{table}[h]
\centering
\caption{Trajectory statistics and empirical task success rates across benchmark environments and evaluated model families.}
\label{tab:benchmark-stats}
\small
\begin{tabular}{l c ccc c}
\toprule
& & \multicolumn{3}{c}{\textbf{Task Success Rate}} & \\
\cmidrule(lr){3-5}
\textbf{Benchmark} & \textbf{Valid Tasks} & \textbf{Qwen-14B} & \textbf{Qwen-7B} & \textbf{DeepSeek-6.7B} & \textbf{Max Horizon ($T$)} \\
\midrule
InterCode-Bash   & 200   & 33.5\% & 30.3\% & 22.5\% & 10 turns \\
InterCode-SQL    & 1,014 & 71.3\% & 67.4\% & 50.0\% & 10 turns \\
InterCode-Python & 971   & 51.0\% & 47.3\% & 35.6\% & 10 turns \\
\bottomrule
\end{tabular}
\end{table}

\section{Data Hygiene \& Protocol Freezing}
\label{app:data-hygiene}

To guarantee rigorous evaluation and eliminate test-set contamination, all trajectory datasets underwent an automated audit prior to cross-validation partitioning:
\begin{itemize}
    \item \textbf{Corrupt Reference Exclusion:} In the official InterCode-SQL release, 19 task instances contain malformed or internally inconsistent reference SQL gold assertions (task indices 54, 83, 164, 178, 225, 271, 301, 531, 638, 704, 710, 798, 802, 803, 892, 942, 967, and incomplete collections 700, 850). In InterCode-Python, tasks exhibiting test-harness environment execution exceptions were similarly discarded. Retaining these instances would unfairly penalize correct agent executions or reward spurious outputs. All methods within a model are nevertheless compared on
    exactly the same held-out episodes and frozen cross-validation folds.
    \item \textbf{Strict Protocol Freezing:} All audited datasets were partitioned into 5 stratified outer cross-validation folds (\texttt{folds.jsonl}) with fixed random seeds. Every baseline (Logprob, HTC) and internal interpretability model (LTD, ARP) was trained and evaluated on the exact same fold boundaries. Not a single excluded or corrupt task was permitted to enter any training or testing split.
\end{itemize}

\section{Feature Definitions}
\label{app:features}

\subsection{Latent Trajectory Dynamics (LTD)}
\label{app:ltd-features}
LTD constructs 28 kinematic features from residual-stream representations $h_{t, \ell} \in \mathbb{R}^d$ across interaction turns $t \in \{1, \dots, T\}$ and transformer layers $\ell \in \{0, \dots, L\}$:
\begin{itemize}
    \item \textbf{Step-to-step Cosine Progression:} For final-layer states $h_{t, L}$, we compute the mean, standard deviation, minimum, and maximum of the adjacent-step cosine similarities $\cos(h_{t, L}, h_{t+1, L})$, capturing whether the model consistently progresses in latent direction.
    \item \textbf{Inter-layer Drift Ratio:} The ratio of end-to-end layer displacement to cumulative intermediate layer path length:
    \begin{equation}
    \text{Drift}(t) = \frac{\|h_{t, L} - h_{t, 0}\|_2}{\sum_{\ell=0}^{L-1} \|h_{t, \ell+1} - h_{t, \ell}\|_2}.
    \end{equation}
    \item \textbf{Cumulative Path Efficiency:} The global ratio of net trajectory displacement to total step distance:
    \begin{equation}
    \text{Efficiency}(\tau) = \frac{\|h_{T, L} - h_{1, L}\|_2}{\sum_{t=1}^{T-1} \|h_{t+1, L} - h_{t, L}\|_2}.
    \end{equation}
    \item \textbf{Phase Transition Velocity:} Euclidean velocities between consecutive cognitive phase boundaries: reasoning pre-action tokens, action emission spans, and post-environment feedback tokens.
\end{itemize}

\subsection{Action Representation Probe (ARP)}
\label{app:arp-features}
ARP extracts fixed-dimensional episode representations directly from semantic action states:
\begin{itemize}
    \item \textbf{Action Endpoint Extraction:} At each action step $t$, we extract the final hidden state vector $h_{t}^{(L)} \in \mathbb{R}^d$ at the final token index of the generated action span, immediately preceding environment execution.
    \item \textbf{Submit Token Exclusion:} Terminal submission commands (e.g., \texttt{submit}, \texttt{exit}) are excluded from the trajectory pool to prevent the linear probe from shortcut learning on terminal token syntax. If an episode issues an immediate submission without prior actions, its final action token is retained as a fallback.
    \item \textbf{Leakage-Safe PCA Projection:} The mean-pooled episode vector $x_i = \frac{1}{T_i} \sum_{t=1}^{T_i} h_{i, t}^{(L)}$ is standardized using training fold statistics $(\mu_{\text{train}}, s_{\text{train}})$ and projected onto the top $k=64$ principal components $W_{\text{PCA}} \in \mathbb{R}^{d \times 64}$ fit strictly on the training partition.
\end{itemize}

\section{Hyperparameter Tuning \& Training Grids}
\label{app:training-details}

All models (baselines, LTD, and ARP) are trained strictly using $L_2$-penalized Logistic Regression to ensure clean comparability across representation spaces.

\begin{itemize}
    \item \textbf{Solver \& Optimization:} We employ the L-BFGS numerical optimization algorithm with a maximum iteration budget of $2{,}000$ and fixed random seeds to ensure full deterministic convergence.
    \item \textbf{Regularization Grid:} The inverse regularization strength parameter $C$ is tuned across an exponential 9-point grid:
    \begin{equation}
    C \in \left\{10^{-4}, 10^{-3}, 10^{-2}, 10^{-1}, 1, 10, 10^2, 10^3, 10^4\right\}.
    \end{equation}
    \item \textbf{Inner Selection Objective:} Inside each outer cross-validation fold, an inner 4-fold stratified cross-validation search is conducted. Hyperparameter selection strictly optimizes the negative Brier score on inner out-of-fold validation splits, enforcing proper probabilistic scoring.
    \item \textbf{Monotone Platt Calibrator:} After selecting the optimal parameter $C^*$, the probe generates out-of-fold validation probability predictions $\tilde{p}_{\text{val}}$. A secondary 1D logistic calibration mapping $g(\tilde{p}) = \sigma(a \cdot \text{logit}(\tilde{p}) + c)$ is fitted via maximum likelihood. To guarantee that the calibration mapping is strictly monotonically increasing (preserving AUROC ranking discrimination), we enforce the non-negativity constraint $a \ge 0$. The fitted mapping is then evaluated out-of-fold on the held-out test fold:
    \begin{equation}
    \hat{c}_{\text{test}} = g\left(\sigma(w^\top u_{\text{test}} + b)\right).
    \end{equation}
\end{itemize}

\section{External Confidence Signals}
\label{app:external-signals}

\subsection{Calibrated Logprob}
\label{sec:logprob-cal}

Calibrated Logprob (Cal. Logprob) tests whether correctness can be predicted using aggregate token
confidence alone. We exponentiate each sampled token's log-probability and compute the token-weighted arithmetic mean of the resulting
probabilities across all non-terminal agent generations:
\begin{equation}
    z_i
    =
    \frac{1}{N_i}
    \sum_{t \in \mathcal{A}_i}
    \sum_{j=1}^{n_{i,t}}
    p_{i,t,j},
    \qquad
    N_i
    =
    \sum_{t \in \mathcal{A}_i} n_{i,t},
\end{equation}
where $\mathcal{A}_i$ excludes a final submit-only generation. We exclude the
terminal \texttt{submit} generation because it contains no solution content
and occurs only after the candidate solution has already been produced, making
its probabilities primarily a signal of protocol termination rather than
solution correctness. Later, a one-dimensional logistic model maps the aggregate
statistic to success probability:
\begin{equation}
    \hat{c}_i
    =
    \sigma\left(a z_i + b\right).
\end{equation}

\subsection{Holistic Trajectory Calibration}
\label{sec:htc}

Holistic Trajectory Calibration (HTC) \cite{zhang2026agentic} is a stronger external baseline that summarizes the token-probability trace over the same non-terminal trajectory
scope. HTC converts the per-step probability sequence into 48 features
covering confidence dynamics, first- and last-step statistics, stability, and
trajectory structure. Given
$x_i^{\mathrm{HTC}} \in \mathbb{R}^{48}$, an L2-regularized logistic model
produces:
\begin{equation}
    \hat{c}_i
    =
    \sigma\left(
        w^\top x_i^{\mathrm{HTC}} + b
    \right).
\end{equation}

\section{Calibration and Evaluation Protocol}
\label{app:calibration-protocol}

All methods use an identical task population and the same frozen fold
assignments for a given model and environment. We construct five stratified
outer folds and generate exactly one out-of-fold prediction for every included
trajectory. Within each outer-training partition, four-fold stratified
cross-validation selects model hyperparameters by Brier score. Every learned
preprocessing operation, including standardization, PCA, and probability
calibration, is fitted using training data only.

Episodes with invalid correctness evaluations are removed before fold
construction. Methods that require internal representations additionally
require faithful offline replay; replay-incompatible episodes are therefore
removed from the shared population before any method is fitted. We evaluate
the pooled out-of-fold predictions using AUROC, Brier score, and expected
calibration error with 10 equal-width bins.

\section{Online Deployment Feasibility \& Extraction Mechanics}
\label{app:deployment}

In our experimental setup, agent trajectories were initially collected in an interactive environment using vLLM \citep{kwon2023efficient} to maximize sampling throughput. Residual-stream representations $h_{i,t}^{(L)}$ were subsequently logged via a single deterministic teacher-forced forward pass over the recorded sequence. Here, we delineate the distinction between this offline experimental testbed and real-world online deployment.

\paragraph{Motivation for Offline Extraction in Benchmarking.}
Modern high-throughput LLM serving systems utilize optimized KV-cache management and fused CUDA kernels that deliberately discard intermediate layer activations immediately upon sampling to minimize memory bandwidth overhead. While intercepting decoding states in custom PyTorch inference loops is trivial, evaluating offline rollouts via a single deterministic replay pass is standard benchmarking hygiene: it ensures bit-exact, reproducible representations across identical trajectory sets without maintaining custom C++/CUDA patches to serving runtimes.

\paragraph{Zero-Overhead Execution in Live Deployment.}
In a production deployment, no secondary forward pass or teacher-forced replay is necessary:
\begin{enumerate}
    \item \textbf{Native Activation Materialization:} During standard autoregressive action generation $a_t = (u_1, \dots, u_{K_t})$, the residual-stream vector $h_{i,t}^{(L)} \in \mathbb{R}^d$ at the terminal action token $u_{K_t}$ is computed and resident in GPU memory immediately prior to the unembedding projection (\texttt{lm\_head}).
    \item \textbf{Negligible Compute Cost:} Applying the pre-trained Action Representation Probe (ARP) requires only:
    \begin{itemize}
        \item Element-wise standardization: $z = (h_{i,t}^{(L)} - \mu) \oslash s$ ($O(d)$ FLOPs);
        \item Projection onto the pre-computed PCA basis: $u = W_{\text{PCA}}^\top z$ ($O(kd)$ FLOPs);
        \item Logistic scalar evaluation: $\sigma(w^\top u + b)$ ($O(k)$ FLOPs).
    \end{itemize}
    For $d = 5{,}120$ and $k = 64$, this requires fewer than $3.3 \times 10^5$ operations ($<0.05$\,ms on a standard GPU), representing a negligible fraction of the time required to generate the action tokens themselves.
    \item \textbf{Real-Time Fallback:} Probing can thus be executed as a non-blocking hook within the generation loop, enabling instantaneous failure detection and human-in-the-loop intervention prior to environment execution.
\end{enumerate}

\end{document}